# DeepStreet: A deep learning powered urban street network generation module

**Mr. Zhou FANG**
PhD Candidate
The Martin Centre for Architectural and Urban Studies
Department of Archtecture, University of Cambridge, Cambridge, United Kingdom, CB2 1PX
Email: zf242@cam.ac.uk

**Mr. Tianren YANG**
PhD Candidate
The Martin Centre for Architectural and Urban Studies
Department of Archtecture, University of Cambridge, Cambridge, United Kingdom, CB2 1PX
Email: ty290@cam.ac.uk

**Dr. Ying JIN**
University Reader
The Martin Centre for Architectural and Urban Studies
Department of Archtecture, University of Cambridge, Cambridge, United Kingdom, CB2 1PX
Email: yj242@cam.ac.uk

**Word Count:**
      Abstract: 250 words
      Text: 3983 words
      References:395 words
      Tables: 2
      Figures: 11
**Total**: 250+4006+395+(2+11-6)*250= 6401 words

*Submitted [27/02/2020]*

*Z. Fang, T. Yang, and Y. Jin*

**ABSTRACT**
In countries experiencing unprecedented waves of urbanization, there is a need for rapid and high-quality urban street design. Our study presents a novel deep learning powered approach, DeepStreet (DS), for automatic street network generation that can be applied to the urban street design with local characteristics. DS is driven by a Convolutional Neural Network (CNN) that enables the interpolation of streets based on the areas of immediate vicinity. Specifically, the CNN is firstly trained to detect, recognize and capture the local features as well as the patterns of the existing street network sourced from the OpenStreetMap. With the trained CNN, DS is able to predict street networks' future expansion patterns within the pre-defined region conditioned on its surrounding street networks. To test the performance of DS, we apply it to an area in and around the Eixample area in the City of Barcelona, a well-known example in the fields of urban and transport planning with iconic grid-like street networks in the centre and irregular road alignments farther afield. The results show that DS can (1) detect and self- cluster different types of complex street patterns in Barcelona; (2) predict both gridiron and irregular street and road networks. DS proves to have a great potential as a novel tool for designers to efficiently design the urban street network that well maintains the consistency across the existing and newly generated urban street network. Furthermore, the generated networks can serve as a benchmark to guide the local plan-making especially in rapidly- developing cities.

## INTRODUCTION

Over half of the world's population lives in urban areas, and this proportion is expected to increase to 68% by 2050. Projections show that urbanization, combined with the overall growth of the world's population, could add another 2.5 billion people to urban areas by 2050, with close to 90% of this increase occurring in Asia and Africa (*1*). Even in developed countries with a shrinking national population, many of their densest cities are still expanding as migrants are attracted to these areas for better job prospects and health care. These future growths will make the already complex urban structures and fabric even more challenging to understand, plan and design.

In these large and complex cities, coordination between different interventions on buildings, streets, roads, public transport and utility infrastructure is difficult, and in order to ensure the cities operate smoothly 24/7, decisions on extension, demolition and other alterations of the urban environment need to be made very rapidly, and the time windows for planning and designing streets and roads are very narrow. Even in the most technologically advanced cities today, such planning and design decisions are made simply and with few alternative options for comparison. This is far from adequate, since the geometric and spatial configurations of streets and roads have to reflect intricate, multi-dimensional functional, aesthetic, social and environmental requirements. As a result of this inadequacy, cities all over the world have greatly suffered from poorly designed and alienating street and road environments that completely forget the rich cultural heritage of their historic areas.

Developing more advanced digital assistance (DA) tools for planning and design is one of the ways that can effectively elevate the productivity of urban designers and engineers (*2*). On the one hand, since the first DA tools were developed, this approach has facilitated better quality urban design including the street and road networks. On the other hand, recent success have suggested that AI (especially deep learning) are able to enhance out analytical, human-inspired, and humanized artificial intelligence (*3*).

Recently, different machine learning powered generative models have been developed to generate synthetic urban street networks. Researchers proposed a Convolutional Neural Network (CNN) based Variational Auto Encoder (VAE) trained on street network images to capture the patterns of urban street network using low-dimensional vectors and generating new street networks by controlling the encoded vectors (*4*). A Generative Adversarial Network (GAN), StreetGAN, has also been developed to generate street networks that maintained the consistency of the training dataset (*5*). However, above methods can only generate different patterns of urban street networks independently. The urban street networks generated can not automatically and intelligently adapt to existing street network. Inspired by deep learning approaches for image completion, our research pushes the frontiers of the urban street generation model further by developing a novel method for urban street and road networks that is context-aware.

Image completion (also known as image inpainting) is an active computer vision research area that aims to help people complete the missing parts in an image in a content-aware way(*6*). Various approaches dedicated to solving this problem have been emerged (*7*). Recently, deep CNN coupled with GANs have been introduced for smoother and more realistic output images(*8–10*). These techniques can and has been used in many applications that can be summarized into two categories: (1) the removal of unwanted content in an image and (2) the generation of plausible content within the missing part of an image.

In the road network design context, urban street and road networks can be converted into map-like, multi-channel images. The area for new design can be effectively defined as the missing part in an image. By adopting the deep CNN coupled with GANs, a method which has proven the generative capabilities in classic image completion, computer algorithms can detect and self-cluster complex street patterns and predict both gridiron and irregular road networks.

In this research, we aim to make computers to understand the patterns of urban street networks and generate new urban streets and roads within pre-defined area. This has given rise to a new generation module named DeepStreet (DS).

Through the underlying neural networks, DS learns from a given area of existing streets and roads and predicts configurations of streets and roads within a pre-defined area that is new to it. The given urban street and road network is sampled and converted into pixelized data units in a form that can be read by DS. DS is then trained to detect, recognize, and capture the various features of the given urban street networks, with refinement of its predictions based on the loss functions defined. Once training is complete, DS is to be used to predict new streets and roads in and around the given road networks, or existing networks that DS has no knowledge of through the training process.

To test the performance of the module in a real city, we apply it in and around Barcelona's Eixample area, which is a world-famous example in the field of urban and transport planning. The Eixample area has an iconic gridiron network, and the areas surrounding it are characterized by irregular streets and roads as a result of its medival history, hilly terrains and requirements of connecting to the regional expressways (*11*). The tests show that DS can (1) detect and self-cluster different types of complex street patterns in Barcelona, and (2) predict both gridiron and irregular road networks. DS thus proves its significant potential as a novel DA tool for designers to respond rapidly with urban street and road configuration designs that take fuller account of the city's historic legacy and functional requirements. Furthermore, the predictions have the potential for serving as both benchmarks for nd alternative designs to human-generated options, especially in situations where a rapid design response is required.

## METHODS AND DATA
### Data structure

The relationship between topography and street patterns has been discovered and highlighted for decades, and this relationship increases in robustness in hilly regions(*12*). In order to consider both terrain info and surrounding urban street patterns when generating the new urban street networks, two pixelized DS channels (i.e., Road network Channel and Topographic info Channel) are overlayed and set up to form map-like, multi-channel images.

The existing urban street network is stroked, pixelized, and stored in the Road network Channel. The pixels inside the road coverage are then assigned different values which represent the different hierarchy of urban streets. Void areas are assigned value 255 (representing white). To provide adequate terrain information to DS, a topographic map is first converted to raster images, and these are then pixelated and stored in the Topographic info Channel. The value assigned to each pixel is adherent to the following rule: the pixel with the lowest elevation is assigned value 255 (representing highest color intensity); the pixel with the highest elevation is assigned value 0 (representing lowest color intensity); and all other pixels are assigned the values following a linear scale. **Figure 1** illustrates the data structure adopted to develop DS.

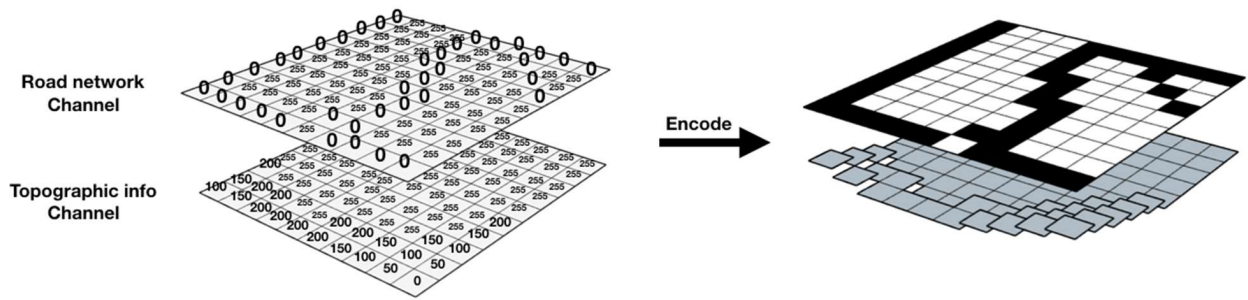

**Figure 1 Data structure for DS**

Based on the specific model requirements and available data sources, **Table 1** summarizes the data items, data sources, and data format.

**TABLE 1 Data sourses and formats**

| Item | Source | Format |
|------|--------|--------|
| Road network | OpenStreetMap (roads layer + railways layer) | Vectors with attributes (level of the road) |
| Topographic Info | USGS/NASA SRTM data (SRTM 90m DEM Digital Elevation Database, http://srtm.csi.cgiar.org/) | Raster images (approximately 30 meters by 30 meters) |

In order to understand the performance of the proposed DS, we applied the data structure proposed to an area in and around Eixample within the City of Barcelona. The urban street map (in a shapefile format) of Barcelona was downloaded from the OpenStreetMap website and a python script was developed to convert the vectors of streets to pixels in images (jpg format). To visualise different road hierarchy, we split the Road network Channel into two RGB channels and the colors used to stroke the polylines (representing roads) are set to three sets of RGB combinations: (255,0,0), (0,255,0) and (0,0,0), which occupies 2 Red and Green channels (**Figure 2**). To emphasize the road hierarchical information, the line width is determined based on the hierarchy level of the road (achieved from the attributes), its actual width (which can be measured and estimated from the Google Maps) and the scale to re-draw the map. In this case study, the size of each pixel represents 5 ×5m.

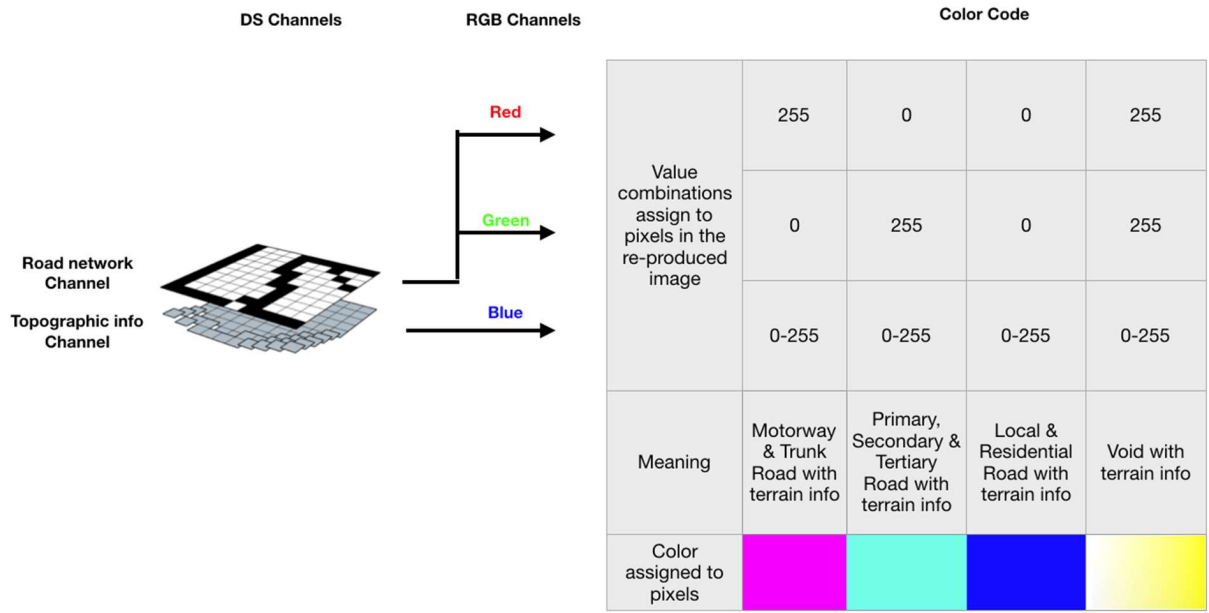

**Figure 2 Value combinations and color codes adopted for the Barcelona Case**

We convert the National Aeronautics and Space Administration (NASA) Shuttle Radar Topography Mission (SRTM) 90m data for the city of Barcelona and stored it into the Topographic info Channe (for visualization, Blue channel). As the SRTM 90m data is stored in a coarse pixelated raster image (where each pixel represents 30 × 30m area), each pixel in the raster image is split into 36 (30m/5m × 30m/5m = 36) smaller pixels (represent 5 × 5m area) and the values representing elevations are rescaled to 0-255 (represent 511 to 0 m a.s.l. in the City of Barcelona case). The reproduced map for the City of Barcelona represented by the Road network Channel and Topographic info Channel using the color codes summarized in **Figure 2** is shown in **Figure 3**.

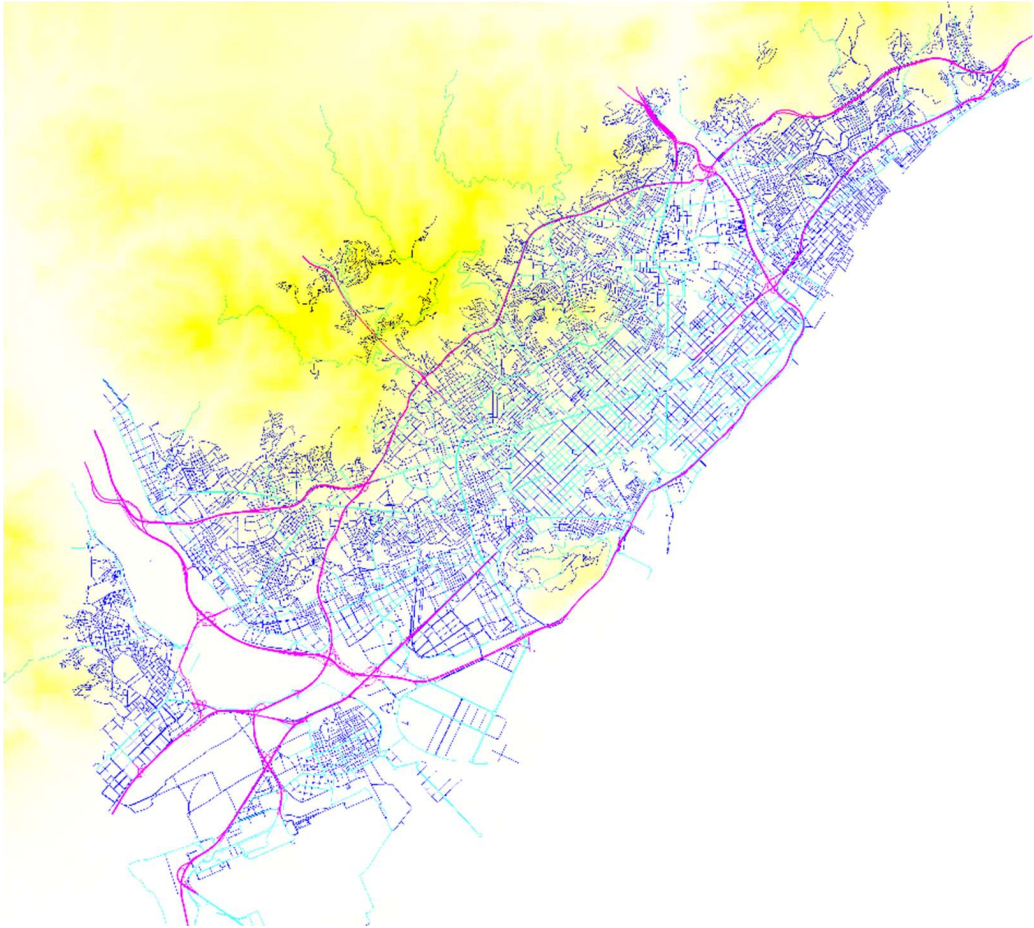

**Figure 3 Reproduced multi-channel image for the City of Barcelona**

**Data collection**

To guarantee the independence and a full coverage of the data units collected, we randomly generated 900,000 unoverlapped unit squares (256 × 256 pixels, represent 1280 × 1280m each; including three RGB channels) located in the prepared multi-channel image (**Figure 3**) and cropped them out for sampling. We then split them into two datasets with 80% for training and 20% for testing. Three representative unit squares showing different types of street patterns are presented in **Figure 4**.

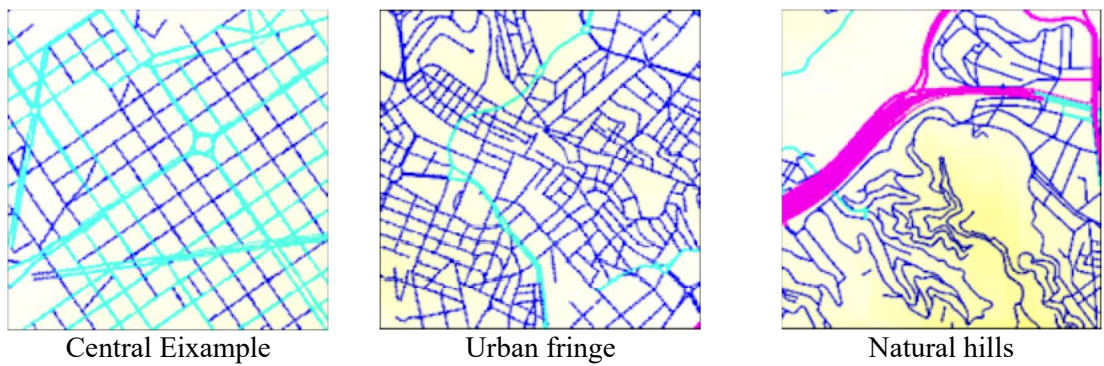

Central Eixample          Urban fringe          Natural hills

**Figure 4 Three types of street patterns in the City of Barcelona**

**Mask generation**

To train the generative capability of the proposed DS, a randomly generated Mask Channel (MC) is set up to define the coverage of the new development areas (regions to generate new urban street network), which is then concatenated to the collected data units before being fed into the DS module. MC shares the same size (256 × 256 pixels) as other channels and only stores 0 and 1 for each pixel in the channel. MC takes the value 0 inside the new development areas and 1 elsewhere. After element-wise multiply MC and other channels (except for Topographic info Channel), the road network information stored within new development areas is cleared and the channels are ready to be used to predict the urban street network. Indeed, the information preserved in the Topographic info Channel can be used to guide the prediction of urban street network in the missing region.

The strategy we adopted to generate MC was one where we first randomly ascertained a center location of the hole in the 64 ×192 pixels range, with the width and height of the hole being 48 pixels (**Figure 5**). This mask generation method guarantees that each pixel in the randomly generated hole can be calculated based on the present of a sufficient amount of context pixels.

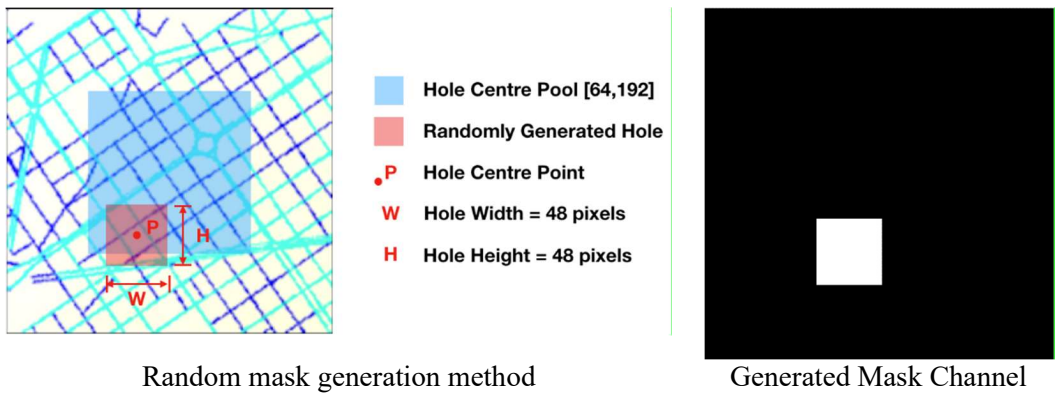

Random mask generation method          Generated Mask Channel

**Figure 5 Mask generation strategy**

**Overall process to set up DS - Training and Testing**

The overall process used to set up DS is shown in **Figure 6**. In the Training stage, Training Set data are fed into the proposed deep neural network with initial parameters batch by batch to recursively calibrate the parameters in the neural network through the use of backpropagation algorithm(*13*). Once the loss defined in the Loss Function cannot be further minimized with further iterations, the DS model is trained. After that, Testing Set data are used to assess the performance of the trained DS module. Similar to the image completion task, DS's performance is difficult to be assessed based on the quantitative evaluation metrics due to the existence of multiple possible solutions (*14*, *15*). Consequently, only the qualitative analysis can be conducted to assess the performance of DS.

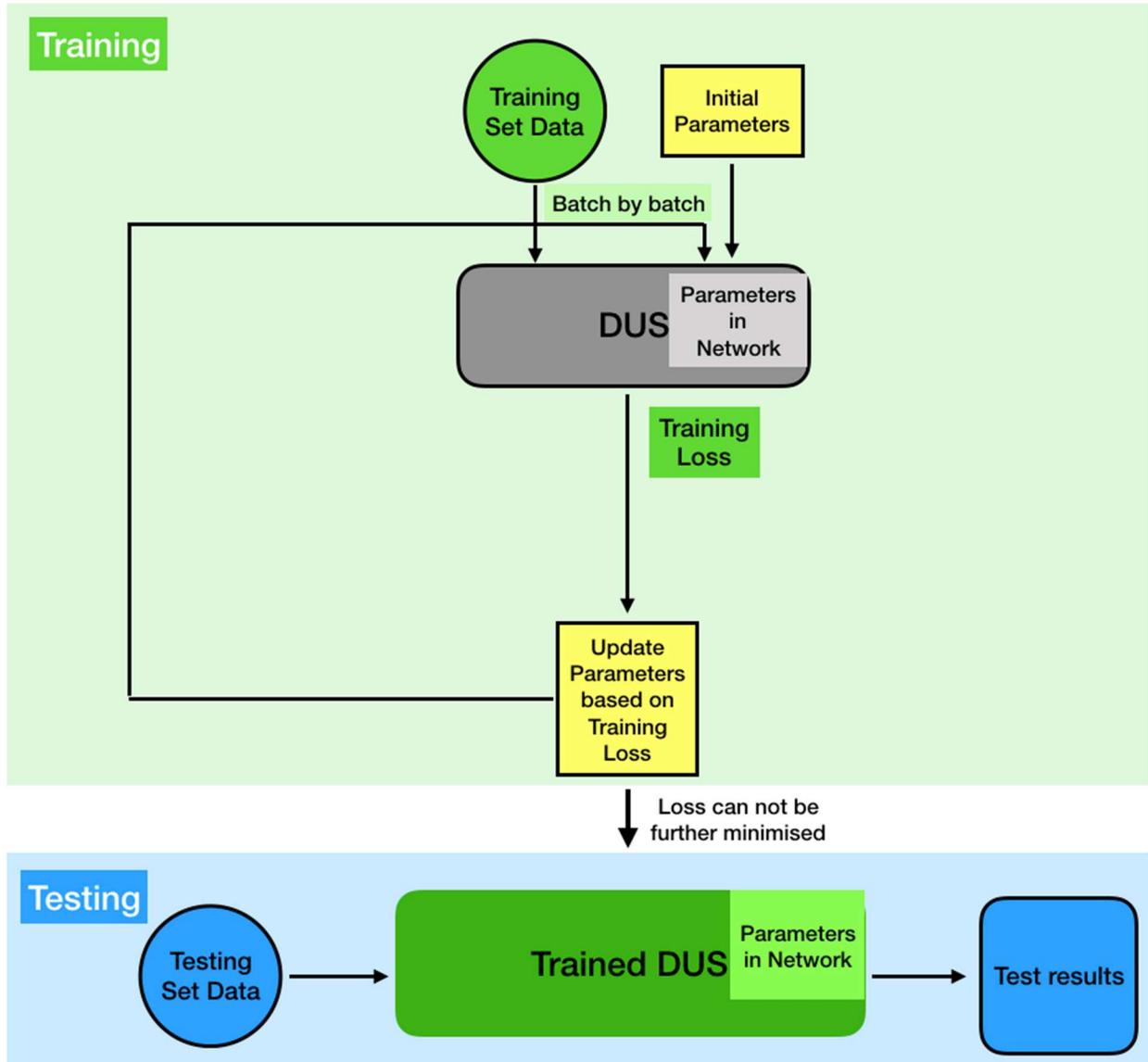

**Figure 6 Overall process to setup DS**

**Proposed deep neural network structure**

The deep neural network proposed for training the DS model is designed based on the network developed for the image completion task. At this stage, our network structure was developed from the structure that has been introduced in the Globally and Locally Consistent Image Completion Method (*16*). Our network structure is made up of two types of networks: a single generation network and two context discriminator networks. The generation network is designed to generate urban street network within a pre-defined area by the Mask Channel. Two context discriminator networks, i.e., the global discriminator and local discriminator, work as GANs to guide the training process towards a more realistic urban street network design. An overview of our network architecture is shown in **Figure 7**.

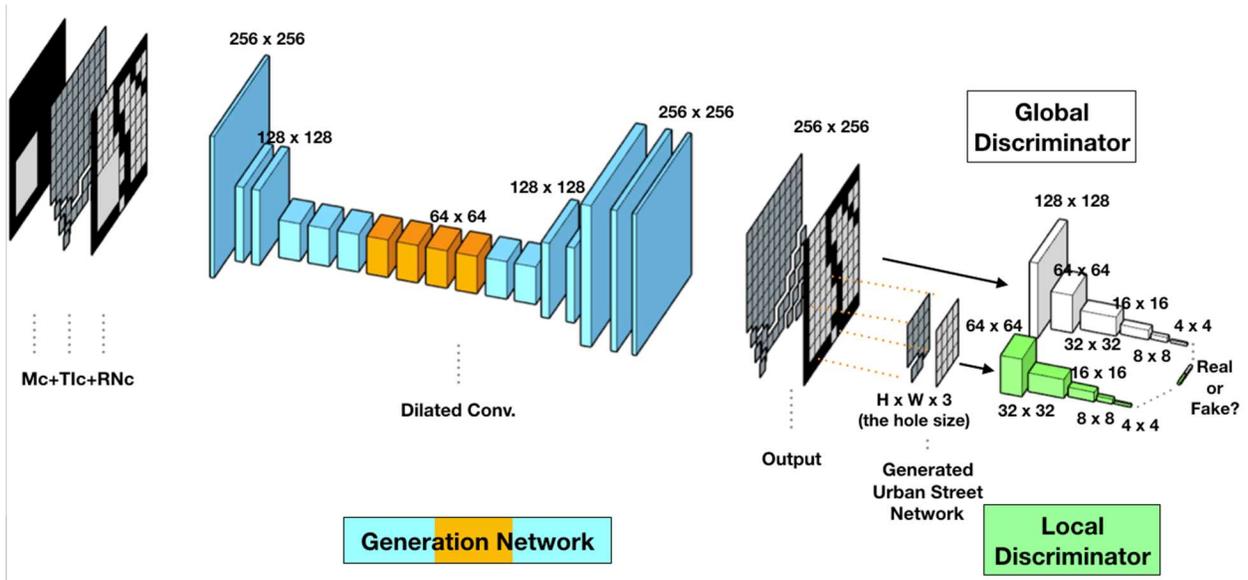

**Figure 7 Overview of the network architecture for DS**

*Generation network*

The generation network is based on a fully convolutional network. The input of the generation network is a Road network Channel (2 RGB channels), a Topographic info Channel (1 RGB channel), and a concatenated Mask Channel. The output is a 256 × 256 × 3 data unit, which is made up by Road network Channel (2 RGB channels) and Topographic info Channel (1 RGB channels). As we did not want any changes in areas other than the pre-defined development areas, the output pixels outside of the development areas were restored to the input values.

The general architecture follows a classic encoder-decoder structure, which allows for an extracting feature in the encoder stage, transforming the context features to predicted features in the fully connected layer, and being restored to the original resolution in decoder stage using the deconvolution layers(*17*). As the spatial relationships among features are crucial to our urban fabric generation task, the pooling layer, which can reduce image resolution by making the spatial relativity between features less relevant, was not used (*10*). In our structure, only the stride convolutions are used to decrease the resolution of the input data unit (as shown in Figure 5, 256 × 256 to 64 × 64). Through this approach, the spatial relationships among features remain and the computational complexity can be significantly reduced.

Dilated convolutional layers are also used in the mid-layers (highlighted in orange in Figure 5) (*18*). Dilated convolutions allow for the computing of each output pixel with a much larger input area, while still using the same amount of parameters and computational power. This is important for our task, as most of the designers would like to design new urban street network with a deeper understating of the surrounding environment.

*Context discriminators*

A global context discriminator network and a local context discriminator network are attached to the generation network to discern whether an image is real (original) or has been generated (completed) from both global and local points of view. By using both global and local context discriminators, the generated urban street network can be trained towards realistic urban texture. Moreover, the generated urban street network can maintain global and local consistency, and this would mean maintaining quality and character of an urban space, without any threat to urban identity and livability.

**Loss function and optimization algorithm adopted**

Before introducing the loss functions adopted for DS training, the terminology used are listed in **Table 2**.

**TABLE 2 Terminology for loss function**

| Terminology | Note | Size |
|---|---|---|
| x | Road network Channel & Topographic info Channel | $256 \times 256 \times 2 + 256 \times 256 \times 1$ |
| $M_c$ | Mask Channel (0 – within the generation region; 1 – elsewhere) | $256 \times 256 \times 1$ |
| $C(x, M_c)$ | The generation network in a functional form. | 17 layers of generation network |
| $D(x, M_c)$ | The combined context discriminators (Global and local discriminators) in a functional form. | The output is in the [0, 1] range and represents the probability that the urban fabric is real, rather than generated. |
| $L(x, Mc)$ | The Mean Squared Error (MSE) loss, which is defined as $\text{MSE} = \frac{1}{n}\sum_{i=1}^{n}(Y_i - \widehat{Y}_i)$, representing the difference between the generated urban fabric and the real one) | |

In this research, two loss functions are used in tandem to train the DS model, a weighted MSE loss, and a GAN loss (*9*). Using the mixture of the two loss functions allows for the stable training of a high-performance network model and has been used for image completion (first proposed in the CE Method) (*10*).

Weighted MSE loss: The weighted MSE loss is proposed to quantify the difference between the generated road network and the real work road network within the development region (defined by the Mask Channel). The MSE loss is defined as:

$$L(x, M_c) = \|M_c \odot (C(x, M_c) - x)\|^2 \qquad (1)$$

where $\odot$ is the pixel-wise multiplication and $\| \cdot \|$ is the Euclidean norm.

GAN loss: GAN loss is considered and included in the loss functions of most image completion algorithms. By taking GAN loss into account, the standard MSE loss minimization process can be turned into a min-max optimization problem in which the discriminator networks are jointly updated with the completion network at each iteration. This is crucial for our approach, due to the existence of multiple possible solutions. In our network, the optimization is:

$$\min_{C} \max_{D} \text{E}[log D(x, M_c) + \log{(1 - D(C(x, M_c), M_c))}] \qquad (2)$$

By combining the two loss functions (MSE and GAN), the optimization becomes:

$$\min_{C} \max_{D} \mathrm{E}[L(x, M_c) + \alpha log D(x, M_c) + \alpha \log (1 - D(C(x, M_c), M_c))] \qquad (3)$$

where α is a weighing hyper-parameter.

For optimization, the ADADELTA algorithm is used, as this can automatically set a learning rate for each weight in the network (*19*).

## RESULTS
### Settings for model training and training time

In this research, DS is trained using 720,000 unrepeated data units (Training Set), which are randomly cropped from the multi-channel, map-like image, with a small missing part. We set the weighting hyper-parameter to α = 0.001, and train using a batch size of 24 data units. The generation network is trained for 900,000 iterations (=30 epoch), the discriminator is trained for 30,000 iterations (=1 epoch), and finally, joint networks are trained for 900,000 (=30 epochs). The entire training procedure takes roughly 180 hours (7.5 days) to complete on a desktop equipped with a single NVIDIA RTX 2080Ti GPU.

### Computational time

The processing time of a single urban street network generation task ($256 \times 256 \times 3$ pixels input) using a trained DS is around 0.211 sec (on a desktop equipped with single NVIDIA RTX 2080Ti GPU).

### Successful cases

For the urban street network generation task, the results that come out are generally acceptable, as can be seen by some of the selected results for analysis shown in **Figure 8** (more results shown in **Figure 11**). It is easy to recognize that the patterns of generated urban street networks are consistent with the patterns in their context region. Moreover, all the lines generated are successfully connected to the network in the context region with corrected line widths and colors. This means that (1) during the training stage, DS can detect various complex street patterns and understand complex spatial relationships. (2) the trained DS can detect road patterns in the context region and generate similar road network (same type of pattern) within the missing region, (3) the trained DS can thus adapt/connect generated road networks to the context network.

From the T1 and T2 cases, as shown in **Figure 8**, we can notice that the trained DS performs well when generating the road network within central Eixample. This is because the features of the road network therin are simple and have a strong coherence.

From the T3 to T6 cases, we can see that although all the results are different from the ground truths, the results themselves are natural and realistic (difficult for people to discern whether the road network is an actual one or a generated one). Considering DS generates road networks purely relying on the surrounding road networks, all these results can be seen as acceptable and welcome ones at this stage. Once other types of information (constraints), such as built form, land use, etc., are incorporated into DS, the road networks generated will be more similar to the ground truths.

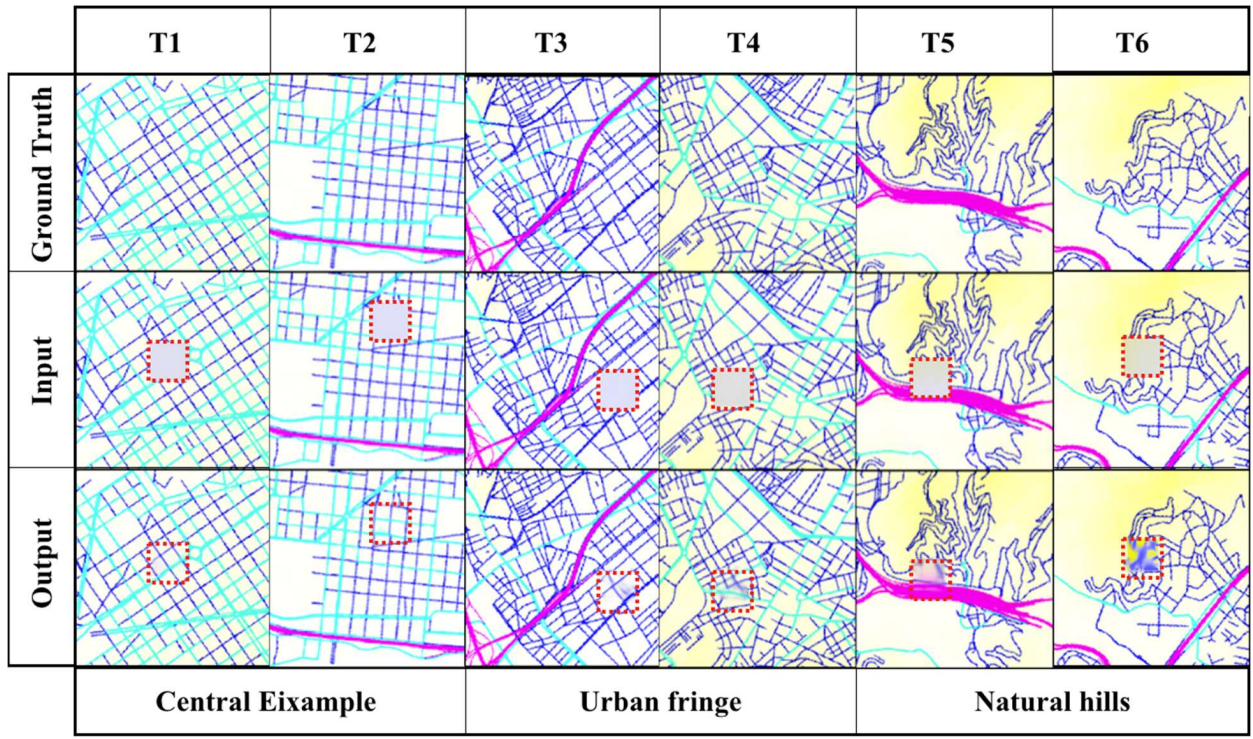

**Figure 8 Selected successful cases for the case study**

**Failure cases**

In the results, some failure cases are noticed. Selected cases of failure are shown in **Figure 9**. The outputs have shown that the trained DS cannot successfully generate lines (either straight or curve) by connecting the ends of the lines at the boundary of the missing region. This type of failure frequently occurs when the missing region is surrounded by a large blank area in the Road network Channel. The reason is as follows: each pixel inside the missing region is calculated based on the surrounding pixels. If the surrounding pixels are all share the same value, 255, in the Road network Channel, the calculated values of pixels inside the missing part have a high probability of being 255. To overcome above problem identified, more information can be provided in the missing region to guide the DS during the generation process.

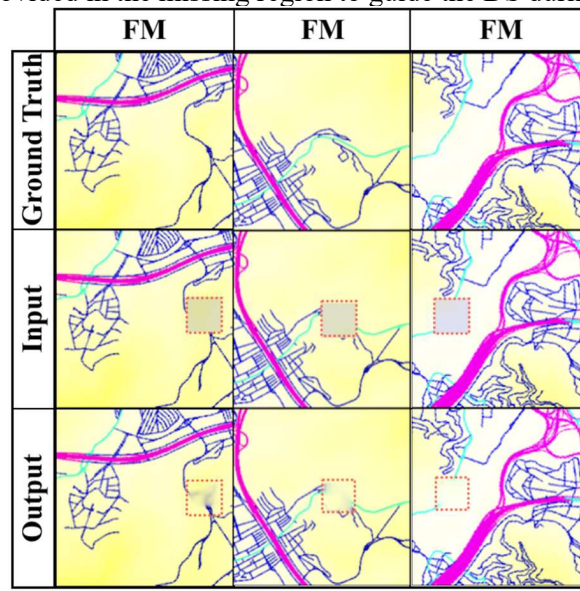

**Figure 9 Examples of Failure Mode (FM)**

**DISCUSSION AND CONCLUSION**

In summary, the proposed DS performed as expected, and most of the results were encouraging and promising. More importantly, the case study for Barcelona city helped us to better understand the capability of DS and inspired us with failed cases to further improve the DS module.

From the case study, we found that DS has the advantages of:

a. Sense/Perception. DS can detect and self-clustering different types of complex road patterns, including the grid-like pattern in the central Eixample, mixed patterns in the urban fringe and winding patterns in natural hills.

b. Generative capability. DS can perform well especially when generating simple and coherent road network such as gridirons. However, such capability decreases when it comes to irregular road network e.g., in natural hills.

At the same time, the limitations of DS are clear and are summarized as follows:

a. Limited predictive capability. DS cannot generate road patterns that are non-existent in the context region.

b. Limited creativity. DS is unable to create new urban street patterns for landmarks, iconic, and special transport arrangement, etc. without designer's interventions.

With the co-existence of the limitation and the capabilities the DS module demonstrated, we have identified opportunities to improve the DS module. First, we clearly observed that urban street network and topographic info alone cannot accurately predict the pattern of urban street network. Building layout, land use types, and other urban elements also interactively influence the urban street patterns and urban fabrics(*20*). To imitate the ways that designers and engineers designing the urban space, we are extending the DS module to incorporate more types of urban elements into the data units, including: building footprint, land use and activity types, etc. So far, we have proposed a new data structure, as illustrated in **Figure 10**. By doing so, the DS module proposed in this research will be upgraded to a DeepUrbanSpace (DUS) module, which can help urban designers to generate urban designs by simultaneously consider various type of information. This can provide a novel approach to delivering an integrated urban design. Moreover, the upgraded DUS module can help us to insightfully understand the urban physical fabric (i.e., streets and built form) and the relationships among different urban elements. In addition, we have proposed a Designer-guided Channel to store and transmit the designers' design idea to computer modules(*15*). In the Designer-guided Channel, some simple sketches are stored, and urban designers can use this channel and these sketches to introduce landmarks, preferred street patterns and transport hub locations into the DUS. As such, through the use of this additional information, the DUS can generate urban fabric that would have more relevance and higher desirability.

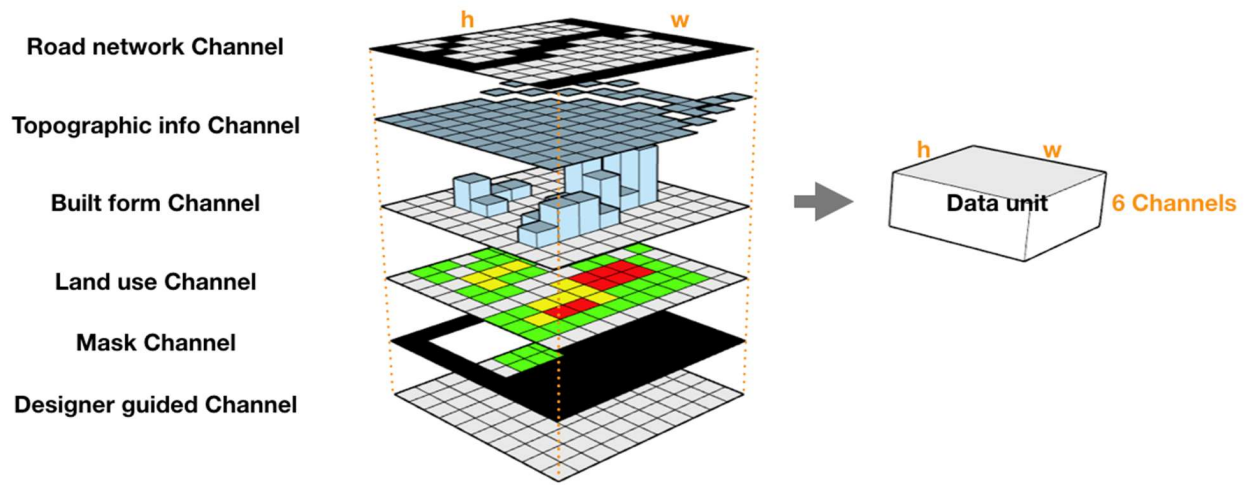

**Figure 10 Structure of the data units for DUS**

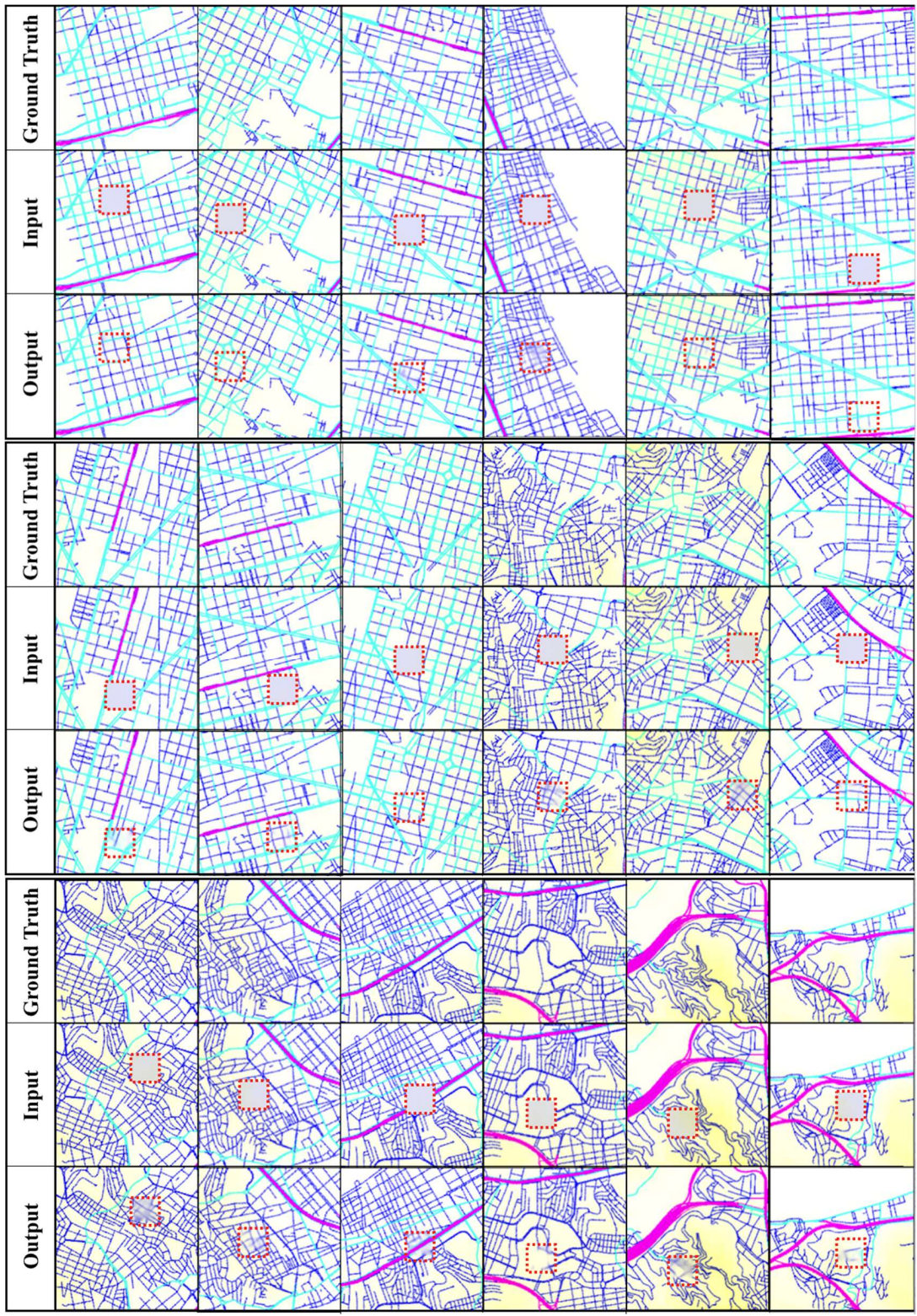

**Figure 11 Additional results generated by DS using randomly generated masks**

**AUTHOR CONTRIBUTIONS**

The authors confirm contribution to the paper as follows: study conception and design: Z. Fang; data collection: Z. Fang; analysis and interpretation of results: Z. Fang, T. Yang; draft manuscript preparation: Z. Fang, T. Yang; Research task supervision: Ying Jin. All authors reviewed the results and approved the final version of the manuscript